\documentclass[runningheads]{llncs}
\usepackage[T1]{fontenc}
\usepackage{booktabs}
\usepackage[misc]{ifsym}

\usepackage{bbding}

\usepackage{amsmath}
\usepackage{float}
\usepackage{xspace}
\usepackage{hyperref} 
\usepackage{subfig}

\usepackage{graphicx}
\usepackage{algorithm}
\usepackage{algorithmic}

\usepackage{xcolor}
\usepackage{hyperref} 
\usepackage{float}
\usepackage{xspace}
\usepackage{subfig}

\usepackage{color}

\begin{document}
\title{Revisit Time Series Classification Benchmark: The Impact of Temporal Information for Classification}
\author{Yunrui Zhang \Envelope \and
Gustavo Batista \and
Salil S. Kanhere}
\institute{University of New South Wales, Sydney NSW 2052, Australia\\
\email{\{yunrui.zhang, g.batista, salil.kanhere\}@unsw.edu.au}}

\authorrunning{Y. Zhang et al.}
\titlerunning{Revisit Time Series Classification}
%

%
\maketitle              
\begin{abstract}
Time series classification is usually regarded as a distinct task from tabular data classification due to the importance of temporal information. However, in this paper, by performing permutation tests that disrupt temporal information on the UCR time series classification archive, the most widely used benchmark for time series classification, we identify a significant proportion of datasets where temporal information has little to no impact on classification. Many of these datasets are tabular in nature or rely mainly on tabular features, leading to potentially biased evaluations of time series classifiers focused on temporal information. To address this, we propose UCR Augmented, a benchmark based on the UCR time series classification archive designed to evaluate classifiers' ability to extract and utilize temporal information. Testing classifiers from seven categories on this benchmark revealed notable shifts in performance rankings. Some previously overlooked approaches perform well, while others see their performance decline significantly when temporal information is crucial. UCR Augmented provides a more robust framework for assessing time series classifiers, ensuring fairer evaluations. Our code is available at \url{https://github.com/YunruiZhang/Revisit-Time-Series-Classification-Benchmark}.

\keywords{Time Series Classfication  \and Temporal Information.}

\end{abstract}
\section{Introduction}

Classification is one of the most successful tasks in machine learning, making it unsurprising that time series classification is also among the most extensively researched topics in time series analysis. Each year, dozens of methods are proposed, and hundreds of application papers are published. However, considering that most time series classification methods focus on univariate and fixed-length time series, the question arises: What distinguishes time series classification problems from their tabular counterparts? We argue that time series classifiers should ideally be capable of exploiting the correlations or patterns between data points while also tolerating misalignment across different instances to improve classification accuracy.

Most time series classification methods uncover temporal information by extracting relevant subsequences interpreted as ``shapes'' or ``features''. This includes many state-of-the-art techniques such as the shapelet transform classifier~\cite{STC2}, the random convolutional kernels classifier~\cite{dempster_etal_2021}, and feature extraction approaches such as Catch22~\cite{catch22}. The proposed methods are benchmarked on many time series classification datasets, usually from the UCR time series classification archive~\cite{UCRArchive}. Nonetheless, despite a strong emphasis on extracting and utilizing temporal information for time series classification, limited research has focused on the significance of temporal information as a contributing factor.


To address this gap, we propose a test to evaluate the contribution of temporal information in time series classification datasets by disrupting their temporal structure. By comparing the performance of time series classifiers on the original and disrupted datasets, we find that removing temporal information does not significantly impact classification for a large portion of datasets in the UCR archive. This is primarily because these datasets are tabular in nature or too well-segmented, reducing the reliance on temporal information for classification.

The observation that temporal information is not a critical feature for classification in these datasets suggests that the current benchmark is sub-optimal for evaluating a classifier's ability to extract and utilize temporal information—the main focus of time series classification. To address this, we propose a new benchmark UCR Augmented, based on the UCR time series classification archive \cite{UCRArchive}, designed to emphasize the importance of temporal information and overcome the limitations of the original archive.

Using this benchmark, we evaluate seven categories of time series classifiers representing diverse approaches and analyze their performance. Our findings reveal notable shifts in the relative rankings of these classifiers, with some previously overlooked methods performing strongly under the new benchmark. In contrast, tabular and phase-dependent methods that performed well on the original benchmark show significantly weaker performance on the proposed benchmark. The contributions of this paper can be summarized as follows:

\begin{enumerate}
\item Proposing a test for identifying datasets where temporal information does not significantly contribute to classification.
\item Identifying a significant proportion of the UCR archive datasets where temporal information has minimal impact and analyzing the cause.
\item Introducing a new benchmark that evaluates the classifier's ability to extract and utilize the temporal information.
\item Benchmarking major categories of time series classifiers on the proposed benchmark and analyzing changes in their relative performance compared to the UCR archive to provide insights.
\vspace{-10pt}
\end{enumerate}

\section{Related Work}\label{sec:bg}
\vspace{-5pt}
Time series classification assigns discrete labels to data sequences, with temporal order in varying indexes being crucial features, setting it apart from tabular data classification. The UCR time series classification archive \cite{UCRArchive} has significantly contributed to the field's growth by providing a standard benchmark. This archive includes 128 heterogeneous univariate time series datasets from various application domains, with 112 having equal sequence lengths.

The current research in time series classification typically introduces new methods and assesses their performance using the UCR time series classification archive \cite{UCRArchive}. This includes shape-based methods that employ different techniques for extracting and comparing shapes for classification \cite{STC2} and feature-based approaches that convert time series data into features\cite{catch22}. All these methods emphasize the extraction of temporal information from time series data. However, no prior work has examined the importance of temporal information for classifying time series in the benchmark archive. 

Bagnall et al. \cite{BakeoffStanardClassifiers} benchmarked various classifiers on the UCR time series archive, finding that some standard tabular data classifiers outperformed the DTW-1NN algorithm, which accounts for temporal information. Dhariyal et al. \cite{dhariyal2023basics} found that tabular methods could outperform time series classifiers on a subset of the UCR archive. Henderson et al. \cite{henderson2023never} showed that features like mean and standard deviation performed well on specific datasets. These findings, where classifiers ignoring temporal information still achieve good results, made us question the universal importance of temporal information in time series classification. Hu et al. \cite{RealisticAssumptions} shows that many time series datasets are unrealistically well-aligned and lack non-representative subsequences, unlike real-life tasks.
\vspace{-5pt}
\section{Preliminaries}\label{sec:def}
\begin{definition} \textbf{Time Series} A time series is an ordered set of n real-valued variables. For a time series \(X\) it can be defined as \(X = x_{1}, \ldots, x_{n}\).
\end{definition}
\begin{definition}  \textbf{Subsequence} For a time series \(X\) of length \(n\), a subsequence \(S_{z}\) of \(X\) is a sub-interval of \(X\) of length \(m\) that \(m < n\) with the starting position of \(z\). \(S_{z} = x_{z}, \ldots x_{z+m}\) for \(1\leq z \leq n-m+1\)
\end{definition}
\begin{definition} \textbf{Temporal Information} For a time series \(X\) and all possible subsequences, the temporal information \(\mathcal{T}\) for each data point \(s_{t}\) in a subsequence \(S\) is the conditional probability \(p(s_{t} \mid s_{1 \ldots t-1, t+1 \ldots m})\) where $m$ is the length of the subsequence.
\label{def3}
\end{definition}

As discussed in the introduction, time series classification differs from tabular data classification due to the critical role that temporal information plays in classification. As explained in Definition \ref{def3}, temporal information can be seen as the correlation between data points in subsequences, regardless of the starting point of the subsequence. In practice, temporal information is often interpreted as a unique shape or feature of the subsequence, irrespective of its index.

\section{Testing the Impact of Temporal Information} \label{sec:exp}

In this section, we propose the Temporal Information Removal Test to assess whether temporal information is a contributing factor for each time series classfication dataset. The concept of the test is straightforward: it involves transforming the time series instances in the dataset by applying the same permutation to every instance of the dataset to remove temporal information while preserving the information from a tabular perspective. This transformation is invariant for tabular methods or Euclidean distance but disrupts the sequential nature of the data, thereby removing the temporal information.

We benchmark time series classifiers on both the original and transformed datasets. Since time series classifiers rely on subsequence extraction methods, such as intervals or convolution, and the transformation disrupts the sequential nature of the time series thus eliminates the subsequence features, we expect a significant drop in classification accuracy on the transformed datasets compared to their accuracy on the original datasets, assuming temporal information is a major contributing factor. If, however, the classifier's accuracy does not decrease for certain datasets following this data permutation procedure, it suggests that temporal information may not significantly impact classification for those specific datasets.

To remove temporal information, the proposed transformation involves permuting all time series instances across the training and testing sets using the same randomly generated index. For a given time series \(X\) with a sequence length \(n\), a permuted index array \(I_{p}\) is generated by randomly shuffling the original index array \(I_{o} = [1, 2, 3, \ldots, n]\). This \(I_{p}\) is then used to reorder all instances in \(X_{train}\) and \(X_{test}\), effectively permuting every time series instance in both sets using the same \(I_{p}\). This permutation disrupts the sequential nature of the data, thereby destroying the temporal information.

The subsequences of the permuted time series will differ significantly from the original ones and will not retain the same sequential features. While this transformation is expected to negatively impact the performance of time series classifiers, as it removes the temporal information these classifiers rely on, our focus is on datasets where the accuracy does not decrease significantly. This suggests that, for those datasets, the time series classifier did not rely on temporal information for classification in the original dataset.

Two time series classification algorithms are selected to evaluate the original and permuted datasets: Mini-Rocket \cite{dempster_etal_2021} and Catch-22  \cite{catch22}. These classifiers represent two different approaches to time series classification: pattern-based and feature-based. Catch-22 \cite{catch22} is regarded as a leading method in the feature-based category, and Mini-Rocket \cite{dempster_etal_2021} is noted for its strong performance and good efficiency. Both algorithms rely on subsequence features and the sequential nature of the data: Mini-Rocket generates features using convolution, whereas Catch-22 derives most of its feature set from subsequence information or temporal statistics.

The selected classifiers underwent five runs with different train-test splits, different seeds for permutation and different classifier random states for both the original and permuted data. To select datasets where temporal information does not significantly contribute to classification for further investigation, we apply the filtering rule: $Acc_{per} >= Acc_{ori} \text{ OR } \lvert Acc_{ori}-Acc_{per}\rvert <= Std_{ori}+Std_{per}$. The filtering rules select a set of datasets with the mean accuracy across the five runs of the permuted dataset higher than the original dataset or the difference within the sum of standard deviation across the five runs. 

After applying the filtering rule for Mini-Rocket and Catch-22, we found that 38 (34\%) and 36 (32\%) of the 112 UCR datasets, respectively, had original and permuted data accuracy within the sum of standard deviation. Of these, 12 datasets for both Mini-Rocket and Catch-22 showed mean accuracy on permuted data equal to or higher than the original. When relaxing the criteria to two sums of standard deviations, the numbers increase to 55 (49\%) and 56 (50\%) datasets for Mini-Rocket and Catch-22, respectively. This intriguing observation suggests a significant portion of UCR datasets does not rely on temporal information for classification. Additional evidence supports this finding: the tabular data classifier Rotation Forest \cite{Rotation}, which does not consider temporal information, as compared to the state-of-the-art meta-ensemble time series classifier, Hive-COTE 2 (HC2) \cite{middlehurst_hive-cote_2021}. Using the same filtering rule, Rotation Forest had 44 datasets within one standard deviation and 60 within two, with most datasets overlapping with those identified in our experiment.

These results lead us to question what makes these datasets less dependent on temporal information for classification. If temporal information is not the key feature, what features are the classifiers relying on?

\section{Cause Analysis} \label{sec:case}
In this section, we follow up on our previous discovery and provide some insights into why temporal information is not a significant contributing factor for some datasets in the current time series classification benchmark.
\subsection{Perfect Alignment}
One major issue we observed that causes the temporal information not to contribute to classification significantly is that some datasets come from a source with equal sequence lengths and a standardized format that does not require segmentation or has been well segmented and appears perfectly aligned. Many of these datasets have near-perfect alignment, causing distinguishing features to appear in the same position for all instances. This renders the ordering information useless, resulting in tabular data classification methods and time series classifiers performing similarly on the permuted and original data since the temporal information becomes irrelevant.

\begin{figure}[htb]
   \vspace{-20pt}
    \centering
    \subfloat[\scriptsize{Perfect Alignment}]{
        \includegraphics[width = .47\columnwidth]{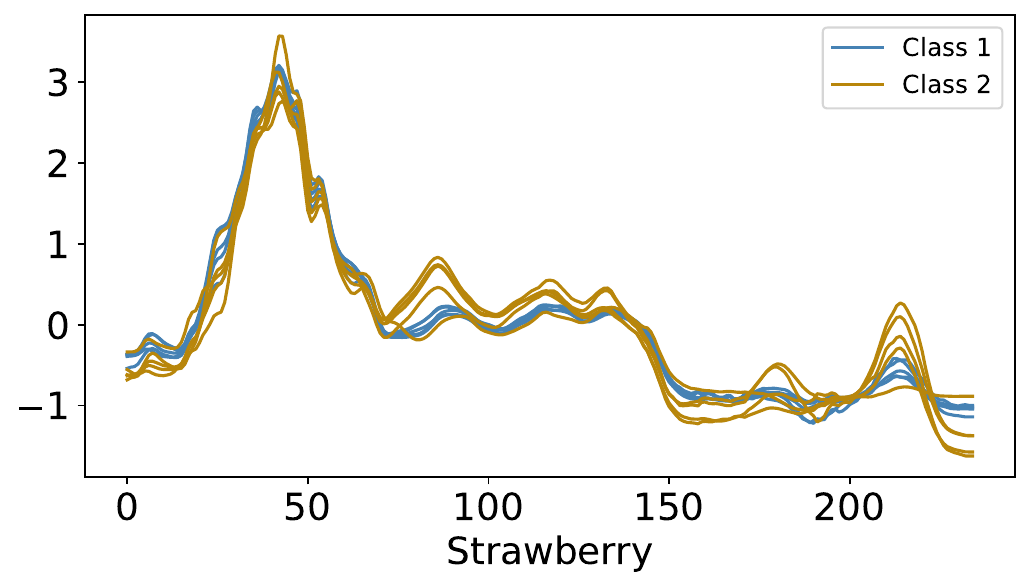} 
        \label{fig:Strawberry}
    }
    \subfloat[\scriptsize{Over segmentation}]{
        \includegraphics[width = .47\columnwidth]{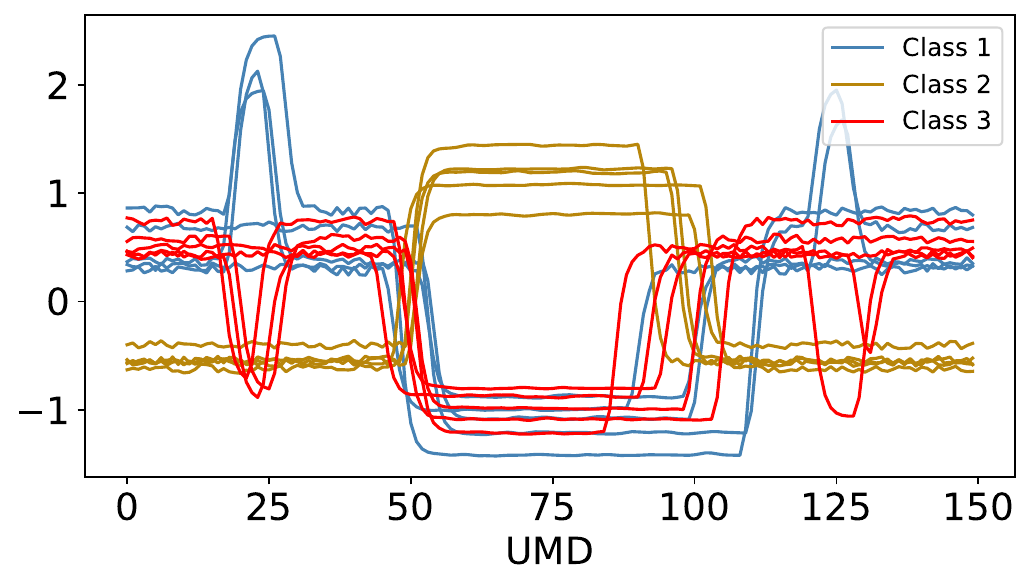} 
        \label{fig:UMD}
    }
     \caption{Example datasets the temporal information is not a significant contributing factor}
    \label{fig:cause}
    \vspace{-20pt}
\end{figure}

Taking the spectrum analysis dataset Strawberry as an example illustrated in Figure \ref{fig:Strawberry}. In this dataset, each instance is a spectroscopic analysis graph, which visually represents how a substance interacts with electromagnetic radiation. The x-axis indicates the wavelength or frequency, while the y-axis represents the intensity or magnitude of the response. The indexes on the x-axis represent the same wavelength value across all instances, rendering this seemingly sequential characteristic irrelevant since each index represents the same attribute across all instances. The dataset can be viewed as a tabular dataset where each index is a feature, making temporal information an insignificant feature for classification. The current UCR archive includes eight datasets that fall into the spectrum analysis category. Additionally, several datasets contain indices representing the same feature across all instances, making them resemble tabular data.

\subsection{Overly Well-Segmented} \label{sec:distortion}
Another problem we observed is that some datasets, although not perfectly aligned as shown in Figure \ref{fig:Strawberry}, are relatively well-aligned, with only slight misalignment in the distinguishing subsequences. Moreover, these datasets typically lack intra-class distortion in their distinguishing subsequences, meaning there is minimal variation in the shape or magnitude of the time series within each class.

Taking the UMD dataset in Figure \ref{fig:UMD} as an example, while the dataset is not perfectly aligned, the distinguishing subsequences are relatively well aligned and lack intra-class distortion. This combination of characteristics makes these problems less representative of time series data and more akin to tabular data, where tabular classifiers could potentially learn the features and generalize effectively with sufficient training data over the slight misalignment in the simple distinguishing features. These datasets contain time series with minimal noise and are well-aligned, largely due to the segmentation process that is conducted manually or guided by domain knowledge \cite{hu_classification_2016}, making temporal information a less significant contributing factor, rendering the dataset similar to tabular data.

\section{UCR Augmented} \label{sec:AUG}

As discussed in the previous section, we identified a significant proportion of datasets in the UCR archive whose temporal information does not significantly contribute to classification. A closer examination of these datasets revealed shared characteristics that make them resemble tabular data, thereby diminishing the role of temporal information. This limitation renders the UCR archive suboptimal for benchmarking classifiers' ability to perform real-time series classification, where the extraction and utilization of temporal information should be the primary focus. This might explain why tabular methods can deliver strong performance on a subset of the UCR archive and even outperform some methods that account for temporal information \cite{bagnall2020rotation,BakeoffStanardClassifiers}. Such results could create a misleading impression and lead to potentially inaccurate benchmarks.

While removing tabular-like datasets from the current benchmark might seem like a straightforward solution to solve this issue, this approach would significantly reduce the number of datasets available for evaluation. Furthermore, the boundary between tabular and time series data is often ambiguous. Although spectrum analysis datasets are clear examples of tabular datasets, many other datasets are less definitive. It is challenging to argue that temporal information has no contribution to classification in these cases; instead, we can only conclude that its contribution is insignificant compared to the features derived from tabular-like properties. 

Another approach to making the current benchmark more temporal informa\-tion-driven and less tabular is to reduce the influence of tabular features while emphasizing the importance of temporal information in these benchmarks, which is a more critical aspect for evaluating time series classifiers. The approach would focus on minimizing the contribution of tabular-like features and designing benchmarks that enable temporal information to be a critical feature for classification. 

To implement this approach, we propose UCR Augmented, a benchmark that augments the UCR time series classification archive \cite{UCRArchive}. The UCR Augmented reduces the influence of tabular features while emphasizing the importance of temporal information by introducing misalignment into the datasets in the UCR archive, thereby reducing their tabular-like characteristics. Misalignment is introduced by adding sequential padding, generated through a Gaussian random walk of different lengths, to both ends of each instance of the z-normalized original univariate time series\footnote{For univariate time series classification, the z-normalization for each instance is recommended \cite{UCRArchive}.}.

The two padding \(Pad_{H}\) and \(Pad_{T}\) of length \(n_{h}\) and \(n_{t}\) are added to the beginning and ending to the original time series respectively, The padding \(Pad_{H}\) and \(Pad_{T}\) are generated in a Gaussian random walk manner, where each element \(Pad_{i}\) is generated $Pad_{i} \sim \mathcal{N}(Pad_{i-1},\,0.01)$. While the $Pad_{0}$ is set to be the \(X_{0}\) and \(X_{n}\) the first and last element of the original time series instance for \(Pad_{H}\) and \(Pad_{T}\) respectively. The padding generated for \(Pad_{H}\) is then taken as the reverse to connect to the beginning of the original time series. 
 
\begin{figure}[htb]
\vspace{-20pt}
    \centering
    \subfloat[\scriptsize{Equal length padding}]{
        \hspace{-10pt}
        \includegraphics[width = .49\columnwidth]{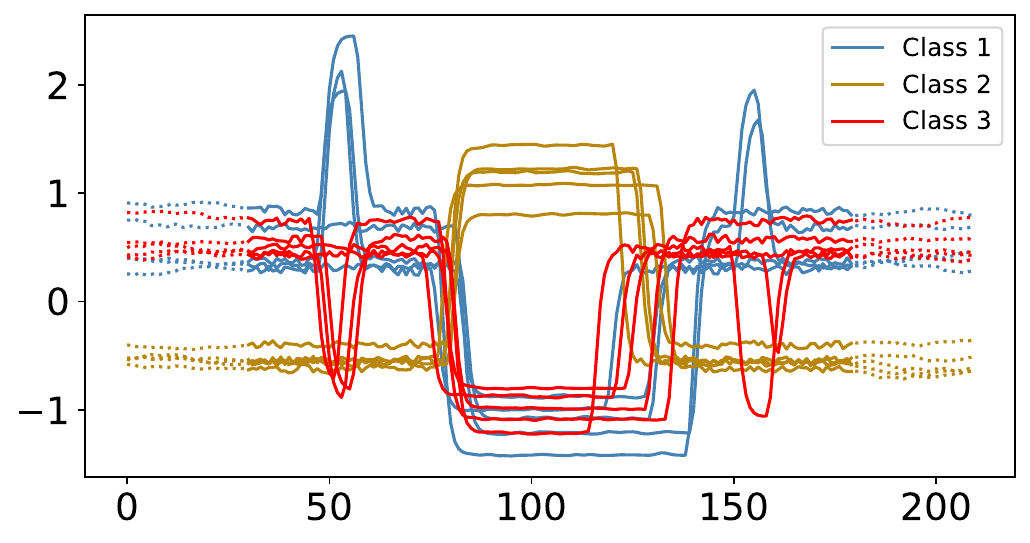} 
        \label{fig:padded}
        \hspace{-5pt}
    }
    \subfloat[\scriptsize{Random length padding}]{
        \includegraphics[width = .49\columnwidth]{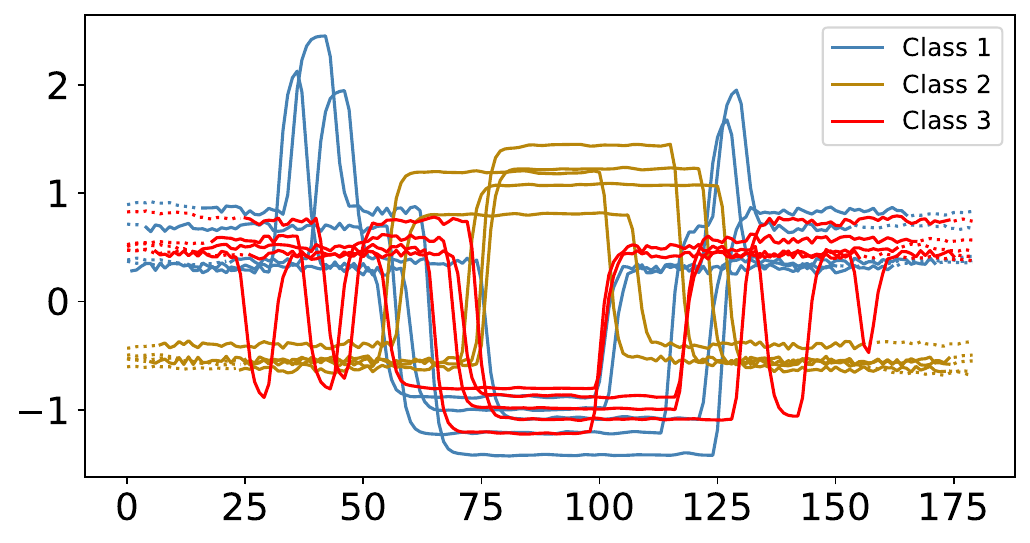} 
        \label{fig:shifted}
    }
     \caption{Example of the UMD dataset with equal-length padding and random-length padding, illustrated with padding shown as dotted lines.}
    \label{fig:aug}
    \vspace{-15pt}
\end{figure}

 As illustrated in Figure \ref{fig:padded}, which shows an example of the dataset UMD with equal-length padding. The Gaussian random walk padding initialized with the end values of the original time series provides a smooth connection to the original time series. Additionally, the Gaussian random noise is generated with a small standard deviation of \(0.01\), ensuring smooth padding that does not introduce tabular or sequential features that could affect the classifier's performance. 
 

To introduce misalignment the padding length \(n_{h}\) and \(n_{t}\) are randomly sampled from uniform distribution and \(n_{h} + n_{t} = l\). Here, \(l\) is a parameter for controlling the total length of the padding to add for the augmentation. The \(n_{h}\) is determined by uniform distribution with an upper bound set to be \(l\). The \(n_{h}\) and \(n_{t}\) are determined as follows $n_{h} \sim \mathcal{U}\{0, l\}$ and $n_{t} = l - n_{h}$. 
 
The random length padding introduces the misalignment needed to remove the tabular features from the time series. An example of the dataset UMD after the augmentation is shown in Figure \ref{fig:shifted} with \(l\) set to be \(20\) percent of the original time series length. The parameter \(l\) controls the level of augmentation applied to the dataset, where a higher \(l\) indicates a higher level of misalignment augmentation.



\section{Benchmark \& Analysis} \label{sec:benchmark}
To analyze the UCR Augmented benchmark and assess how both time series classier of different categories and also tabular classifiers perform when the tabular features become less relevant while the temporal information becomes more important for classification. We conducted benchmark evaluations of seven time series classifiers, each belonging to a distinct category and representing different approaches to time series classification. They are namely:

\begin{description}
    \item[Distance Based] Dynamic Time Wrapping (DTW-1NN) \cite{DTW_OG}
    \item[Dictionary-based] Word Extraction classifier (WEASEL) \cite{WEASEL}
    \item[Feature-based] Canonical Time-series Characteristics (catch22) classifier \cite{catch22,fulcher2017hctsa}
    \item[Interval-based] Canonical Interval Forest classifier (CIF) \cite{middlehurst2020canonical,TSF}
    \item[Shapelet-based] Shapelet Transform classifier (STC) \cite{STC2}
    \item[Kernel-based] Mini-rocket classifier \cite{dempster_etal_2021}
    \item[Tabular classifier] Rotation forest classifier (RF) \cite{Rotation,bagnall2020rotation}
\end{description}

By benchmarking these classifiers, which represent different approaches to time series classification, on the UCR Augmented dataset and comparing their performance, we aim to investigate how various methods perform when tabular features contribute less to classification while temporal information becomes the key feature for classification.

The seven selected classifiers are trained and tested on the original UCR time series classification archive and the UCR Augmented time series\footnote{Seven datasets are excluded from this evaluation as they could not be trained within a reasonable time frame for some time series classifiers.} with five different padding lengths, $l$, ranging from 10\% to 50\% of the original time series length, \(n\). By comparing the performance drop across different levels of augmentation strength, as well as their performance on the original dataset, we aim to gain insights into their ability to extract and utilise temporal information. We expect all classifiers to suffer a performance drop with just \(l=0.1n\) augmentation due to the diminished contribution of tabular features in many datasets. More importantly, we anticipate that tabular methods will experience the largest performance drop, while phase-independent methods, such as STC, will be relatively less affected.

\begin{figure}[htb]
\vspace{-20pt}
    \centering
    \subfloat[\scriptsize{P-value}]{
        \hspace{-5pt}
        \includegraphics[width = .49\columnwidth]{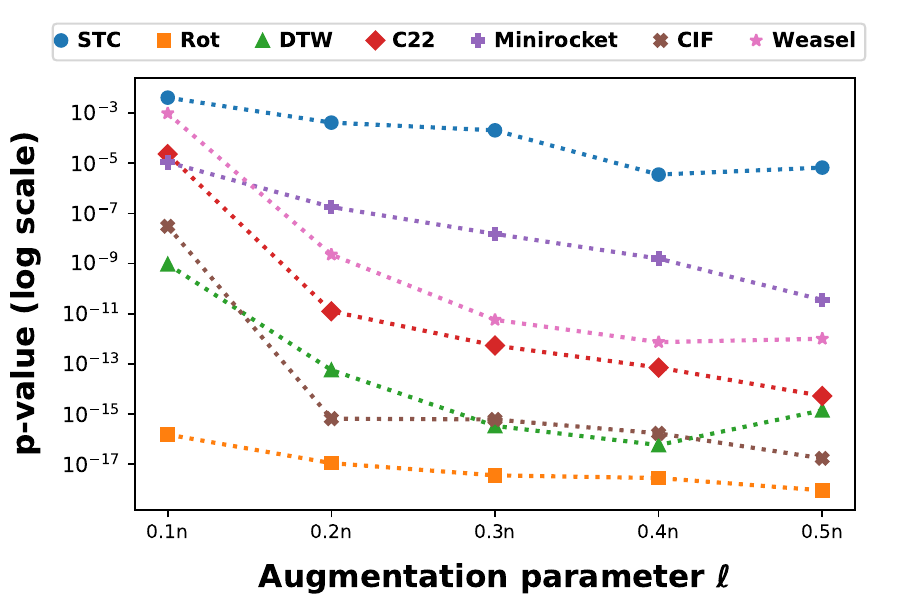} 
        \label{fig:p_val}
    }
    \subfloat[\scriptsize{Mean ranking}]{
        \hspace{-10pt}
        \includegraphics[width = .49\columnwidth]{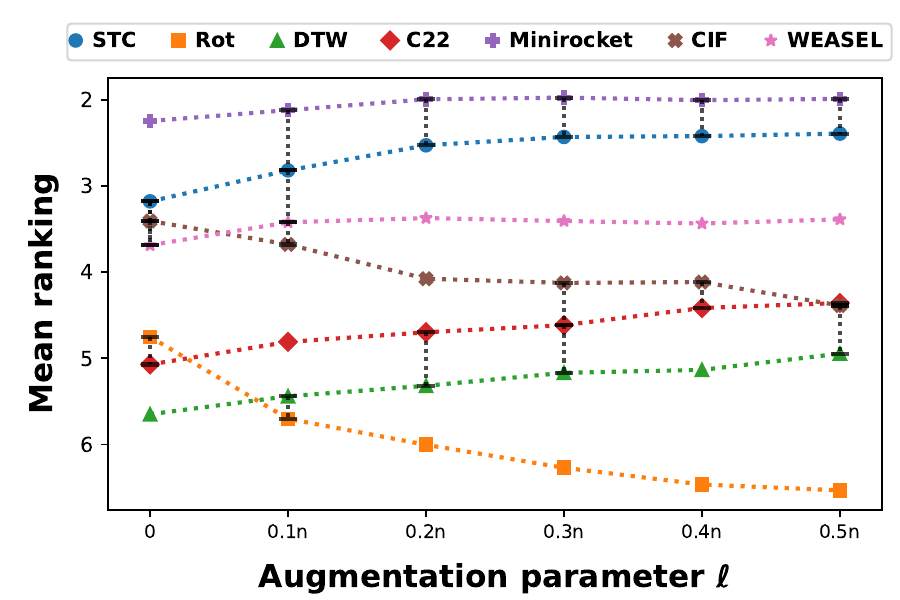} 
        \label{fig:rank}
    }
     \caption{P-value and mean ranking plots of 7 benchmark classifiers on UCR Augmented with \(l\) ranging from \(0-0.5n\). The vertical line in the mean ranking plot indicates no pairwise statistical significance, as tested by the Wilcoxon signed-rank test with a p-value threshold $5 e^{-2}$.}
    \label{fig:aug_trend}
    \vspace{-20pt}
\end{figure}

To evaluate the level of the performance reduction for the classifiers, we performed the Wilcoxon signed-rank test \cite{wilcoxon1992individual} between the classifier's accuracy on the UCR augmented with different $l$  datasets and the accuracy on the original UCR archive datasets, with the alternative hypothesis being the reduction in classifiers' accuracy on the augmented UCR datasets compared to the original UCR archive. As we anticipated, the result reveals that all seven classifiers reject the null hypothesis with the p-value of at least $4.13 e^{-3}$, indicating a significant decrease in accuracy after the slightest augmentation for all the classifiers. 

Figure \ref{fig:p_val} presents the p-values from the Wilcoxon signed-rank test comparing the classifiers' accuracy on the original UCR datasets with their accuracy on UCR Augmented at varying augmentation levels for all selected classifiers. A smaller p-value signifies greater confidence in the alternative hypothesis, indicating the larger performance deterioration for the classifier after the augmentation. Figure \ref{fig:rank} show the mean ranking of each classifier's performance among the seven classifiers on the original UCR time series classification archive datasets and UCR augmented datasets with different \(l\). 
 
\begin{figure}[htb]
 \vspace{-20pt}
    \centering
    \subfloat[\scriptsize{Rotation Forest}]{
        \includegraphics[width = .4\columnwidth]{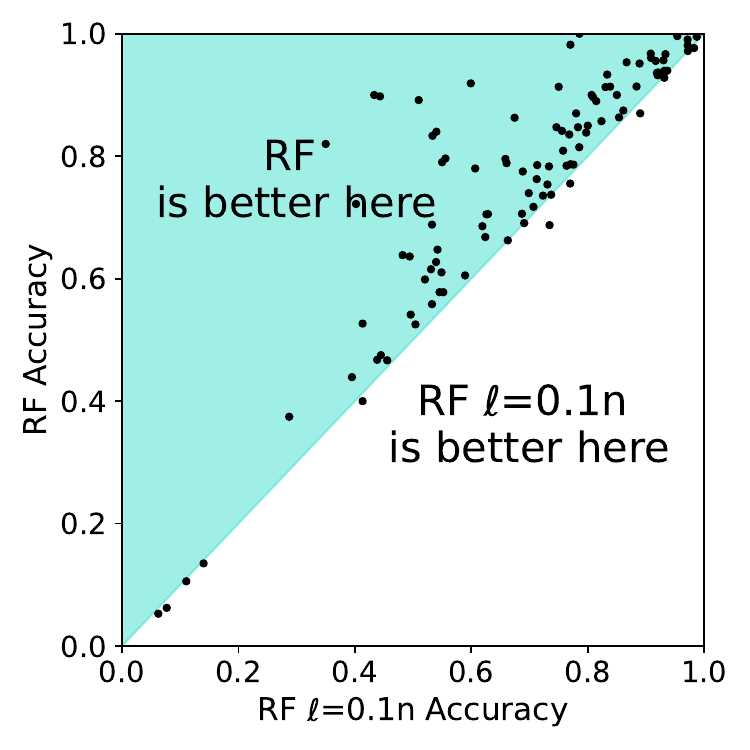} 
        \label{fig:pairwise1}
    }
    \subfloat[\scriptsize{STC}]{
        \includegraphics[width = .4\columnwidth]{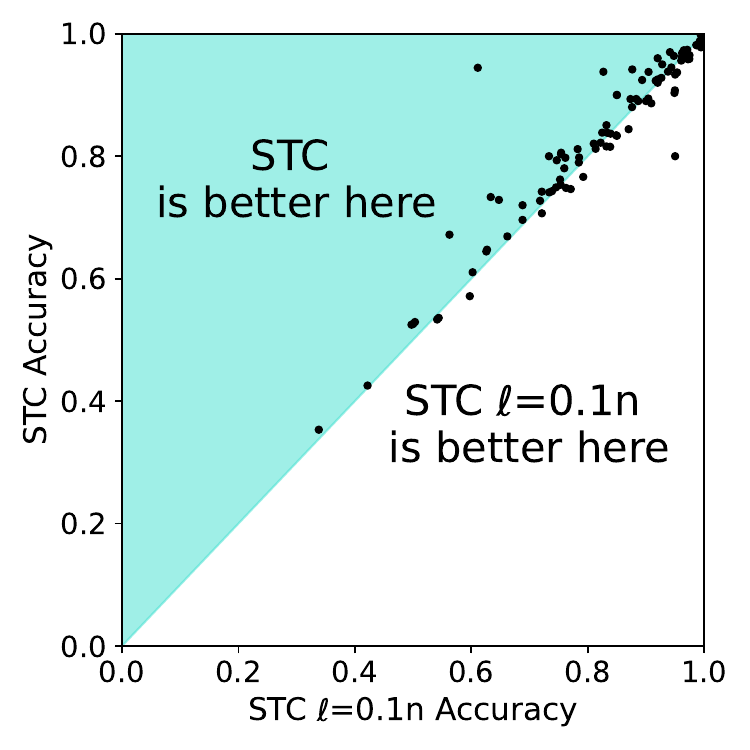} 
        \label{fig:pairwise2}
    }
     \caption{Pairwise Scatter diagrams comparing classifier's accuracy on UCR time series classification archive dataset before and after \(l = 0.1n\) augmentation}
    \label{fig:pairwise}
    \vspace{-20pt}
\end{figure}

As we can observe from Figure \ref{fig:p_val}, the reduction in performance is not equal. As anticipated, the STC suffers much less performance deterioration after the augmentation process than the other classifiers. While tabular method RF suffers more significant performance deterioration, Figures \ref{fig:pairwise1} and \ref{fig:pairwise2} show the pairwise scatter diagram of the accuracy on the original and the augmented UCR time series classification archive datasets that give a detailed visual comparison of the degree of performance deterioration for STC and RF. The unequal performance deterioration also alters the overall ranking of classifiers on the UCR time series classification archive after the augmentation process, as illustrated in Figure \ref{fig:rank}. 

The tabular data classification method RF experienced the greatest performance deterioration and performed significantly worse than all other classifiers, even after the lowest level of augmentation, as shown in Figure \ref{fig:rank}. This result aligns with our expectations, as the augmentation process removed cases where tabular features were key distinguishing factors for classification, making the classification rely more on temporal information. These findings also demonstrate that our augmentation process eliminates tabular features, forcing the model to focus on temporal information for classification. Previous work has pointed out that some tabular classifiers \cite{bagnall2020rotation,BakeoffStanardClassifiers,dhariyal2023basics} exhibit relatively good performance on time series classification benchmark and should serve as a strong baseline or be considered in applications. However, as we have shown, this good performance is often because many datasets in the current benchmark are more tabular in nature than time series. In a time series classification problem where the extraction and utilization of temporal information is the key for classification, tabular methods are not likely to perform well.

Another method that suffers significantly from the augmentation is CIF. As shown in Figure \ref{fig:aug_trend}, CIF overall has the second smallest p-values across all augmentation levels and experiences a large drop in mean ranking with augmentation. The poor performance of CIF can be attributed to the fact that, as an interval-based method, CIF relies on sampling random intervals with the same starting and ending points across all instances of a dataset, making it phase-dependent. This time series classifier assumes that the distinguishing subsequences occur at the same indices across all the cases, an assumption that the augmentation process violates. This strong assumption is not going to hold in real-world time series classification.


Mini-Rocket and WEASEL are two methods that perform relatively well during the augmentation process. These classifiers share a few characteristics: they are all types of ensembles or generate diverse features from the time series. They all utilize sliding windows on the time series instead of global measures on the entire time series. Mini-Rocket uses convolution on the input time series, and many kernels serve as an ensemble. WEASEL uses windowing in various sizes and generates a relatively large number of features before feature selection. The ensemble nature helps these classifiers improve their generalizability and robustness. By operating on segments of the time series either through random intervals or windowing, the distinguishing features from the original time series can still be extracted, which is relatively unaffected by the misalignment.

The STC emerges as the best classifier among the benchmarked classifiers regarding performance deterioration on the UCR augmented, measured by p-value as shown in Figure \ref{fig:p_val}. With the slightest augmentation, the performance difference between the top performers in our benchmark, Mini-Rocket, becomes statistically insignificant compared to STC, as illustrated in Figure \ref{fig:rank}. As a shapelet-based method, the STC relies on the similarity between shapelets and time series as a discriminatory feature. Shapelets are discriminatory subsequences whose quality is assessed and selected by information gain \cite{ye2009time} or hypothesis tests of differences in the distribution of distances between class populations \cite{STC2}. This process is phase-independent; the index of the discriminatory subsequences is not relevant, meaning the degree of misalignment introduced by the augmentation process has little effect on performance. The phase independence characteristic gives shapelet-based methods an inherent advantage in extracting and utilizing temporal information. However, since the shapelet selection process is computationally intensive, shapelet-based methods have received less attention in recent years. The impressive results of STC on the augmented UCR datasets suggest revisiting and further developing these methods, which could be beneficial for future research.

\section{Conclusion} \label{sec:conclusion}

In conclusion, this paper presents novel tests that disrupt the sequential nature of datasets. It reveals that a significant proportion of the datasets in the current benchmark, the UCR time series classification archive, do not rely significantly on temporal information for classification. The primary reason is that many of these datasets are either more tabular than time series in nature or rely predominantly on tabular features for classification.

To better align the current benchmark to evaluate a classifier's ability to extract temporal information, we propose a new benchmark, UCR Augmented, based on the UCR time series classification archive. We evaluated a selected set of classifiers from different categories on this benchmark and provided insights into their performance and changes in ranking compared to the original UCR archive. Our results reveal that some recently overlooked time series classification approaches perform well on the UCR Augmented benchmark. At the same time, tabular and phase-dependent methods experience more significant performance deterioration when temporal information becomes critical for classification. We plan to extend our benchmark for future work by incorporating additional classifiers and expanding the analysis to multivariate time series classification datasets.

\bibliographystyle{splncs04}
\bibliography{references}

\begin{thebibliography}{10}
\providecommand{\url}[1]{\texttt{#1}}
\providecommand{\urlprefix}{URL }
\providecommand{\doi}[1]{https://doi.org/#1}

\bibitem{bagnall2020rotation}
Bagnall, A., Flynn, M., Large, J., Line, J., Bostrom, A., Cawley, G.: Is rotation forest the best classifier for problems with continuous features? (2020)

\bibitem{BakeoffStanardClassifiers}
Bagnall, A., Lines, J., Bostrom, A., Large, J., Keogh, E.: The great time series classification bake off: a review and experimental evaluation of recent algorithmic advances. Data Mining and Knowledge Discovery pp. 606--660 (May 2017)

\bibitem{UCRArchive}
Chen, Y., Keogh, E., Hu, B., Begum, N., Bagnall, A., Mueen, A., Batista, G.: The ucr time series classification archive (July 2015)

\bibitem{dempster_etal_2021}
Dempster, A., Schmidt, D.F., Webb, G.I.: {MiniRocket}: A very fast (almost) deterministic transform for time series classification. In: 27th SIGKDD. ACM (2021)

\bibitem{TSF}
Deng, H., Runger, G., Tuv, E., Vladimir, M.: A time series forest for classification and feature extraction. Information Sciences  \textbf{239},  142--153 (2013)

\bibitem{dhariyal2023basics}
Dhariyal, B., Nguyen, T.L., Ifrim, G.: Back to basics: A sanity check on modern time series classification algorithms (2023)

\bibitem{fulcher2017hctsa}
Fulcher, B.D., Jones, N.S.: hctsa: A computational framework for automated time-series phenotyping using massive feature extraction. Cell systems  \textbf{5} (2017)

\bibitem{henderson2023never}
Henderson, T., Bryant, A.G., Fulcher, B.D.: Never a dull moment: Distributional properties as a baseline for time-series classification. arXiv preprint arXiv:2303.17809  (2023)

\bibitem{STC2}
Hills, J., Lines, J., Baranauskas, E., Mapp, J., Bagnall, A.: Classification of time series by shapelet transformation. DMKD  (Jul 2014)

\bibitem{RealisticAssumptions}
Hu, B., Chen, Y., Keogh, E.: Time series classification under more realistic assumptions. In: SDM, pp. 578--586. SIAM (2013)

\bibitem{hu_classification_2016}
Hu, B., Chen, Y., Keogh, E.: Classification of streaming time series under more realistic assumptions. Data Mining and Knowledge Discovery  \textbf{30} (Mar 2016)

\bibitem{catch22}
Lubba, C.H., Sethi, S.S., Knaute, P., Schultz, S.R., Fulcher, B.D., Jones, N.S.: catch22: Canonical time-series characteristics. DMKD  (Nov 2019)

\bibitem{middlehurst2020canonical}
Middlehurst, M., Large, J., Bagnall, A.: The canonical interval forest (cif) classifier for time series classification. In: 2020 IEEE international conference on big data (2020)

\bibitem{middlehurst_hive-cote_2021}
Middlehurst, M., Large, J., Flynn, M., Lines, J., Bostrom, A., Bagnall, A.: Hive-cote 2.0: a new meta ensemble for time series classification. Machine Learning  (2021)

\bibitem{Rotation}
Rodriguez, J., Kuncheva, L., Alonso, C.: Rotation forest: A new classifier ensemble method. IEEE Transactions on Pattern Analysis and Machine Intelligence  (2006)

\bibitem{DTW_OG}
Sakoe, H., Chiba, S.: Dynamic programming algorithm optimization for spoken word recognition. IEEE Trans on Acoustics, Speech, and Signal Processing  (1978)

\bibitem{WEASEL}
Sch\"{a}fer, P., Leser, U.: Fast and accurate time series classification with weasel. In: CIKM. p. 637–646. ACM (2017)

\bibitem{wilcoxon1992individual}
Wilcoxon, F.: Individual comparisons by ranking methods. In: Breakthroughs in statistics: Methodology and distribution, pp. 196--202. Springer (1992)

\bibitem{ye2009time}
Ye, L., Keogh, E.: Time series shapelets: a new primitive for data mining. In: ACM SIGKDD (2009)

\end{thebibliography}
\end{document}